\pdfoutput=1

\documentclass[11pt]{article}

\usepackage[final]{acl}

\usepackage{times}
\usepackage{latexsym}

\usepackage[T1]{fontenc}

\usepackage[utf8]{inputenc}

\usepackage{microtype}

\usepackage{inconsolata}

\usepackage{graphicx}

\title{RaaS: Reasoning-Aware Attention Sparsity for Efficient LLM Reasoning}

\author{
  \textbf{Junhao Hu\textsuperscript{12*}},
  \textbf{Wenrui Huang\textsuperscript{3}},
  \textbf{Weidong Wang\textsuperscript{3}},
  \textbf{Zhenwen Li\textsuperscript{1}},
  \textbf{Tiancheng Hu\textsuperscript{1}},\\
  \textbf{Zhixia Liu\textsuperscript{4}},
  \textbf{Xusheng Chen\textsuperscript{4}},
  \textbf{Tao Xie\textsuperscript{21$^\dag$}},
  \textbf{Yizhou Shan\textsuperscript{4}} \\
  \textsuperscript{1}SCS, Peking University, Beijing, China\\ 
  \textsuperscript{2}Key Lab of HCST (PKU), MOE, Beijing, China\\
  \textsuperscript{3}School of Computer Science, Nanjing University, Nanjing, China\\ 
  \textsuperscript{4}Huawei Cloud, Shanghai, China \\
}

\usepackage{booktabs}
\usepackage{algorithm}
\usepackage{algorithmic}
\usepackage{amsmath}
\usepackage{amssymb}
\usepackage[skip=0pt]{caption}
\usepackage{pifont}
\usepackage{balance}

\newcommand{\algo}{\textit{RaaS}}
\begin{document}
\maketitle

\renewcommand{\thefootnote}{\fnsymbol{footnote}}
\footnotetext[1]{This work was completed during his internship at Huawei.}
\footnotetext[2]{Corresponding author.}
\renewcommand{\thefootnote}{\arabic{footnote}}
\begin{abstract}
Large Language Models (LLMs) have demonstrated strong capabilities across various domains, with recent advancements in challenging reasoning tasks such as mathematics and programming. However, solving reasoning tasks often requires an LLM to generate long sequences, incurring $O(N)$ time and memory complexities per token, where $N$ is the current sequence length. To reduce complexities, existing sparsity-based algorithms propose to retain Key-Value (KV) vectors, the intermediate representations of only the most critical tokens. However, these algorithms struggle with the ``impossible trinity'' of accuracy, time, and memory. For example, the state-of-the-art algorithm, Quest, achieves high accuracy with $O(L)$ time but $O(N)$ memory ($L$ is the cache budget, $L \ll N$). To address the ``impossible trinity'', in this paper, we identify a new attention pattern during the decode stage of reasoning tasks, where milestone tokens (analogous to lemmas in mathematical proofs) emerge, are utilized, and then become unimportant afterward. Based on this pattern, we propose a new algorithm \algo\ that identifies milestone tokens and retains their KV vectors until they are no longer needed, achieving high accuracy with $O(L)$ time and $O(L)$ memory complexities.
\end{abstract}
\section{Introduction}
\label{sec-intro}

Large Language Models (LLMs) have gained widespread adoption due to their exceptional performance and versatility across various applications. However, their large-scale deployment faces a major obstacle: the high computational cost of long-sequence inference, which is increasingly common in modern user requests (i.e., prompts). This cost arises from the $O(N)$ time and $O(N)$ memory complexities required to generate each token, where $N$ denotes the current sequence length (i.e., input plus output tokens). Consequently, completing an entire request incurs a total time complexity of $O(N^2)$. For instance, the Llama 3.1 8B model supports sequences up to 128,000 tokens, leading to Job Completion Times (JCT) of several thousand seconds and memory usage up to 16GB per request\footnote{\url{https://huggingface.co/blog/llama31}}.

To study long-sequence inference, prior work divides LLMs' generation process into two stages: prefill and decode. First, in the \textbf{prefill} stage, the model processes the prompt tokens given by users. It computes the Key (K) and Value (V) vectors for all prompt tokens, stores these vectors in the KV cache, and generates the first output token to initiate the decode stage. We collectively refer to prompt or input tokens as prefill tokens. Second, in the \textbf{decode} stage, the model iteratively processes each newly generated token. It computes the KV vectors for the new token, appends these vectors to the KV cache, and generates the next token. This process repeats until a specified stopping criterion is met. We refer to output tokens as decode tokens.

\begin{figure*}[t]

\begin{center}
\centerline{\includegraphics[width=\linewidth]{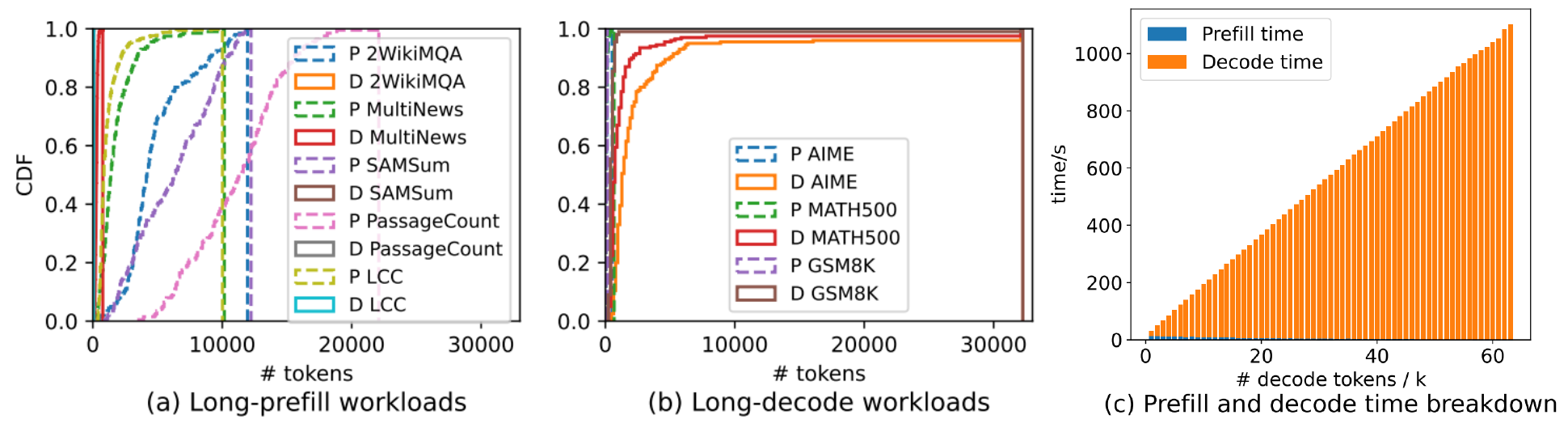}}
\caption{The Cumulative Distribution Function (CDF) of sequence lengths for the Prefill (P) and Decode (D) stages for (a) five datasets from LongBench~\cite{yushi2024longbench} and (b) three math datasets running on the reasoning-enabled Marco-O1 model. (c) The breakdown of prefill and decode time during the inference of fixed 32k tokens using vLLM 0.6.1 with the LLaMA 3.1 8B model in FP16 precision. As the number of decode tokens increases (with the number of prefill tokens being 32k minus the decode tokens), the decode time rises significantly faster than the prefill time.}
\label{fig-background-pd}
\end{center}
\vskip -0.2in
\end{figure*}

This paper focuses on optimizing the decode stage for two main reasons. First, long-decode tasks (producing long model outputs) have recently gained prominence, particularly in reasoning applications, as demonstrated by models such as OpenAI's o1/o3~\cite{openai2024o1} and DeepSeek R1~\cite{damai2024deepseek}. Despite their growing importance, the optimization of long-decode tasks remains under-explored compared to long-prefill tasks (containing long prompts)~\cite{cunchen2024memserve, lianmin2024sglang, woosuk2023vllm, jin2024ragcache, yushi2024longbench}, such as Retrieval-Augmented Generation (RAG), few-shot learning, and tool use. Second, the decode stage represents a significant performance bottleneck in long-decode tasks. In reasoning applications, for instance, the time spent in the decode stage accounts for 99\% of the JCT (Figure~\ref{fig-background-pd}).

Existing sparsity-based algorithms~\cite{tang2024quest, zhang2023h2o, xiao2023sink} optimize long-decode inference by retaining the KV vectors of only the most critical tokens, but struggle with the ``impossible trinity'' of accuracy, time, and memory (Figures~\ref{fig-background-approaches} (b)(c)(d)). First, H2O, the pioneering work on attention sparsity, theoretically achieves $O(L)$ time and memory complexities, where $L$ indicates cache size and $L \ll N$. However, its inability to utilize efficient attention kernels and the lack of page-level KV management make it impractical. Moreover, H2O suffers from low accuracy. Second, StreamingLLM or Sink~\cite{xiao2023sink} similarly offers $O(L)$ time and memory complexities, but adopts an aggressive sparsification strategy that leads to extremely low accuracy on reasoning and other tasks~\cite{tang2024quest}). Third, Quest~\cite{tang2024quest}, the state-of-the-art, achieves high accuracy with $O(L)$ time complexity but $O(N)$ memory complexity.

To maintain accuracy and $O(L)$ time and memory complexities at the same time for reasoning tasks, we analyze their attention pattern during the decode stage, uncovering two key characteristics. First, we identify \textbf{milestone tokens}, which initially exhibit high attention scores but gradually receive lower scores and never receive high scores again. Analogous to lemmas in mathematical proofs, milestone tokens emerge, are utilized, and then fade away. These tokens, visible as bright columns (on the attention map) that slowly diminish, must be carefully managed to prevent significant accuracy loss (Figure~\ref{fig-eval-e2e}). Second, we identify \textbf{phoenix tokens}, which receive low attention scores for a period long enough to be evicted from the cache but later regain importance. These tokens typically appear in the prefill tokens, such as user queries. Quest~\cite{tang2024quest} retains the entire KV cache to avoid losing phoenix tokens, resulting in its $O(N)$ memory complexity.

Based on the preceding observations, we propose a new \algo\ algorithm that addresses the ``impossible trinity'' and consists of two main ideas. First, we identify milestone tokens and retain their KV vectors using a Least-Recently-Used (LRU) caching strategy. During each decoding step, tokens that receive attention scores above the median are considered \textbf{used} and are assigned the latest timestamp. Milestone tokens typically continue to receive the latest timestamps until they become permanently irrelevant. When the KV cache is full, \algo\ evicts KV vectors of the tokens with the oldest timestamp. Second, we retain the KV vectors of all prefill tokens without eviction. Since the phoenix tokens almost always appear within them in reasoning tasks, retaining these tokens' KV vectors ensures that critical information is not lost during the decode stage.

We implement \algo\ with 2k lines of Python code. To evaluate its performance, we compare it against H2O~\cite{zhang2023h2o}, StreamingLLM~\cite{xiao2023sink}, and Quest~\cite{tang2024quest} using three mathematical datasets on four reasoning-enabled models. Our experimental results demonstrate that \algo\ achieves comparable accuracy and latency to Quest, while offering a significant advantage in memory efficiency ($O(L)$ memory complexity). The code is available at: \href{https://github.com/DerekHJH/epic}{https://github.com/DerekHJH/raas}.

In this paper, we make the following three main contributions: 

\begin{itemize}
    \item We identify a novel attention pattern in reasoning tasks, where milestone tokens (analogous to mathematical lemmas) emerge, are utilized, and then become unimportant.
    \item Based on the milestone pattern, we propose a new algorithm \algo\ that achieves high accuracy with $O(L)$ time and $O(L)$ memory complexities. 
    \item We implement and evaluate \algo, demonstrating constant memory usage while maintaining similar accuracy and time performance compared to the state-of-the-art Quest.
\end{itemize}
\section{Background and Motivation}
\label{sec-background}

In this section, we overview the Large Language Model (LLM) inference, highlighting the key concepts and challenges that motivate our work.

\subsection{Autoregressive Generation and KV Cache}
The generation process of LLMs consists of two distinct stages: the prefill stage and the decode stage~\cite{hu2025deepflow, woosuk2023vllm}.
In the \textbf{prefill} stage, the model processes a sequence of prompt tokens all at once. It computes the Key (K) and Value (V) vectors for all prompt tokens, stores these vectors in the KV cache, and generates the first output token to initiate the decode stage.
In the \textbf{decode} stage, the model iteratively processes each newly generated token. It computes the KV vectors for the new token, appends these vectors to the KV cache, and generates the next token. This process repeats until a specified stopping criterion is met. The KV cache~\cite{pope2023efficiently, hu2025epic} accelerates the decode stage by allowing LLMs to process only the new token instead of reprocessing the entire sequence. With the KV cache, the attention mechanism incurs a time complexity of $O(N)$ per decoding step and a memory complexity of $O(N)$ for storing the KV cache, where $N$ is the sequence length.

\subsection{Cost Transfer: From Long-Prefill to Long-Decode Inference}


Long-sequence inference incurs significant costs due to both memory and time requirements. First, it demands substantial memory resources, reaching up to 16 GB KV cache (in addition to the 16 GB model parameters) for processing 128k tokens running the LLaMA 3.1 8B model in FP16 precision\footnote{\url{https://huggingface.co/blog/llama31}}. Second, it requires considerable processing time, with inference for 32k tokens taking around 20 - 1000 seconds on vLLM 0.6.1 using the same model (Figure~\ref{fig-background-pd} (c)).

Long-sequence inference can be categorized into two types: long prefill and long decode. \textbf{Long prefill} arises from extensive input prompts, as observed in prior studies such as Retrieval-Augmented Generation (RAG)~\cite{li2022survey,jin2024ragcache,gao2023retrieval,jeong2024adaptive,ram2023context,mao2020generation} (Figure~\ref{fig-background-pd} (a)). \textbf{Long decode} occurs particularly in reasoning-intensive tasks. Recent advancements emphasize reasoning, where models are guided to think, introspect, and iteratively refine their outputs~\cite{openai2024o1, wang2024openr, lightman2023let, zhao2024marco, wei2022chain}. This approach significantly enhances accuracy but shifts the computational burden to the decode stage. For instance, the OpenAI o1 model~\cite{openai2024o1} requires approximately tens or hundreds of seconds\footnote{\url{https://www.reddit.com/r/OpenAI/comments/1frdwqk/your_longest_thinking_time_gpt4_o1_o1mini/}} of ``thinking time'' before producing its final output. Given the prolonged decoding time and its already substantial proportion of the overall inference process (Figure~\ref{fig-background-pd} (b)), it is critically important to further optimize the decode stage to reduce both time and memory complexities.

\subsection{Existing Sparsity-Based Algorithms}

\begin{figure*}[t]

\begin{center}
\centerline{\includegraphics[width=\textwidth]{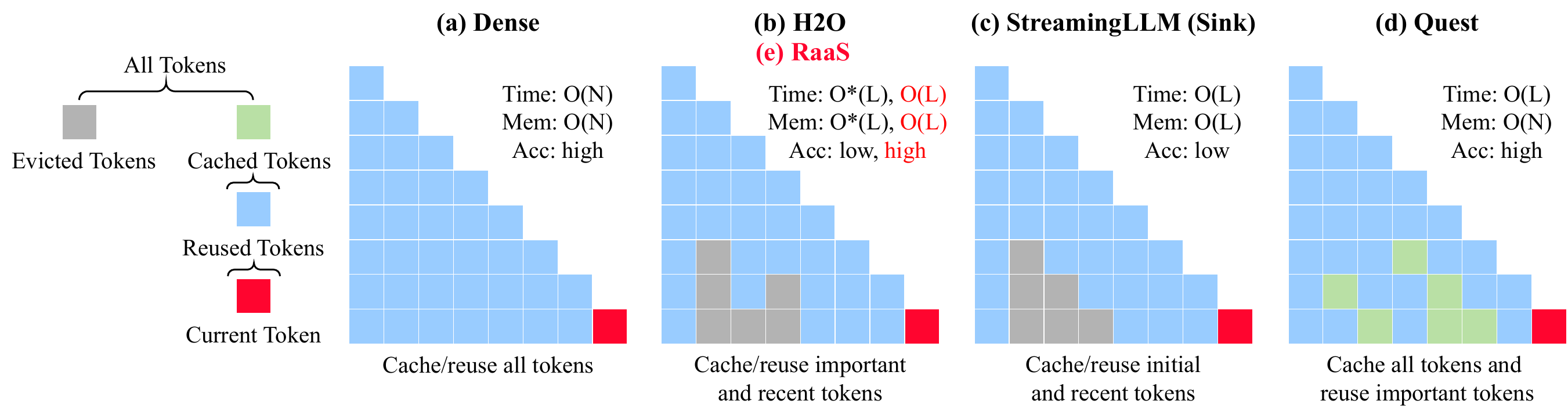}}
\caption{Comparison of sparsity-based algorithms. $N$ indicates the sequence length while $L$ indicates the cache budget where $L\ll N$. Asterisks on H2O's time and memory complexities indicate theoretical complexities that are not realized in practical implementations. \algo\ addresses the ``impossible trinity'' by achieving $O(L)$ complexity for both time and memory, with accuracy comparable to Dense on reasoning tasks. Refer to Section~\ref{sec-background} for detailed explanations of each algorithm's design.}
\label{fig-background-approaches}
\end{center}
\vskip -0.2in
\end{figure*}

To reduce time and memory complexities of long-decode inference, one line of research uses sparsity-based algorithms~\cite{xiao2023sink, zhang2023h2o, tang2024quest, chenarkvale} that retain the KV vectors of only the most critical tokens (fewer than 10\%~\cite{tang2024quest}). But these algorithms struggle with the ``impossible trinity'' of accuracy, time, and memory (Figure~\ref{fig-background-approaches} (b)(c)(d)).

Figure~\ref{fig-background-approaches} compares existing sparsity-based algorithms. First, the Dense or the standard attention algorithm~\cite{vaswani2017attention} caches and reuses KV vectors of all tokens, achieving the highest accuracy but incurring $O(N)$ time and memory complexities. Second, H2O~\cite{zhang2023h2o}, the pioneering sparsity-based algorithm, caches and reuses KV vectors of recent tokens and important non-recent tokens. When the cache is full, it evicts non-recent tokens with the lowest accumulated attention scores. Although H2O theoretically achieves $O(L)$ time and memory complexities, where $L \ll N$ denotes the cache budget, it suffers from low accuracy. Moreover, its lack of support for efficient attention kernels and page-level KV management limits its practical utility. Third, StreamingLLM or Sink~\cite{xiao2023sink} statically decides to cache and reuse KV vectors of only the initial and recent tokens, without dynamically selecting important ones based on attention scores as H2O. Sink also offers $O(L)$ time and memory complexities, but performs poorly on both reasoning and other tasks~\cite{tang2024quest}. Fourth, Quest caches KV vectors of all tokens but reuses only the ones with the top-k attention scores. Quest achieves high accuracy and $O(L)$ time complexity but retains an $O(N)$ memory complexity due to conservative caching.

\section{Algorithm Design}
\label{sec-algo}

\begin{figure*}[t]

\begin{center}
\centerline{\includegraphics[width=\linewidth]{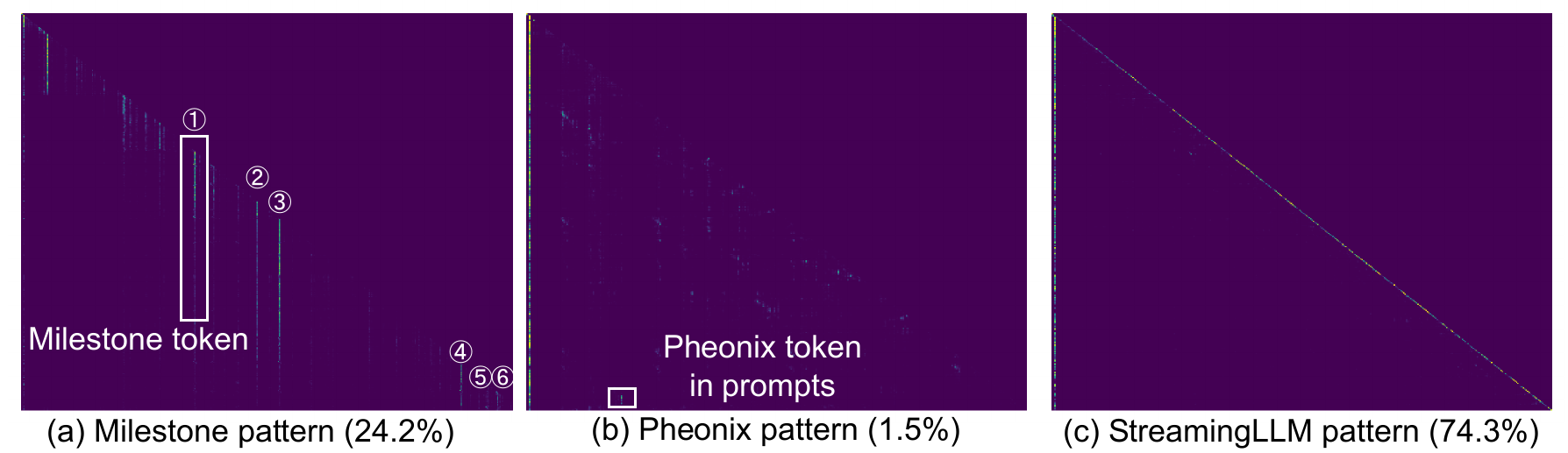}}
\caption{A new attention pattern emerges in reasoning tasks. We manually inspect attention maps across 28 layers and 28 heads of Qwen2.5-Math-7B-Instruct~\cite{qwen25mathtechnical} on 100 MATH500~\cite{math500} test cases. We find that (a) $24.2\%$ maps with milestone tokens, (b) $1.5\%$ maps with phoenix tokens (with a 64-token cache budget), (c) more than 70\% ``lazy''~\cite{zhang2025lighttransfer} maps with StreamingLLM pattern. We use our best effort to balance the clarity and completeness of long-decode attention maps.}
\label{fig-algo-waterfall}
\end{center}
\vskip -0.2in
\end{figure*}

\begin{figure}[t]

\begin{center}
\fbox{\centerline{\includegraphics[width=\columnwidth]{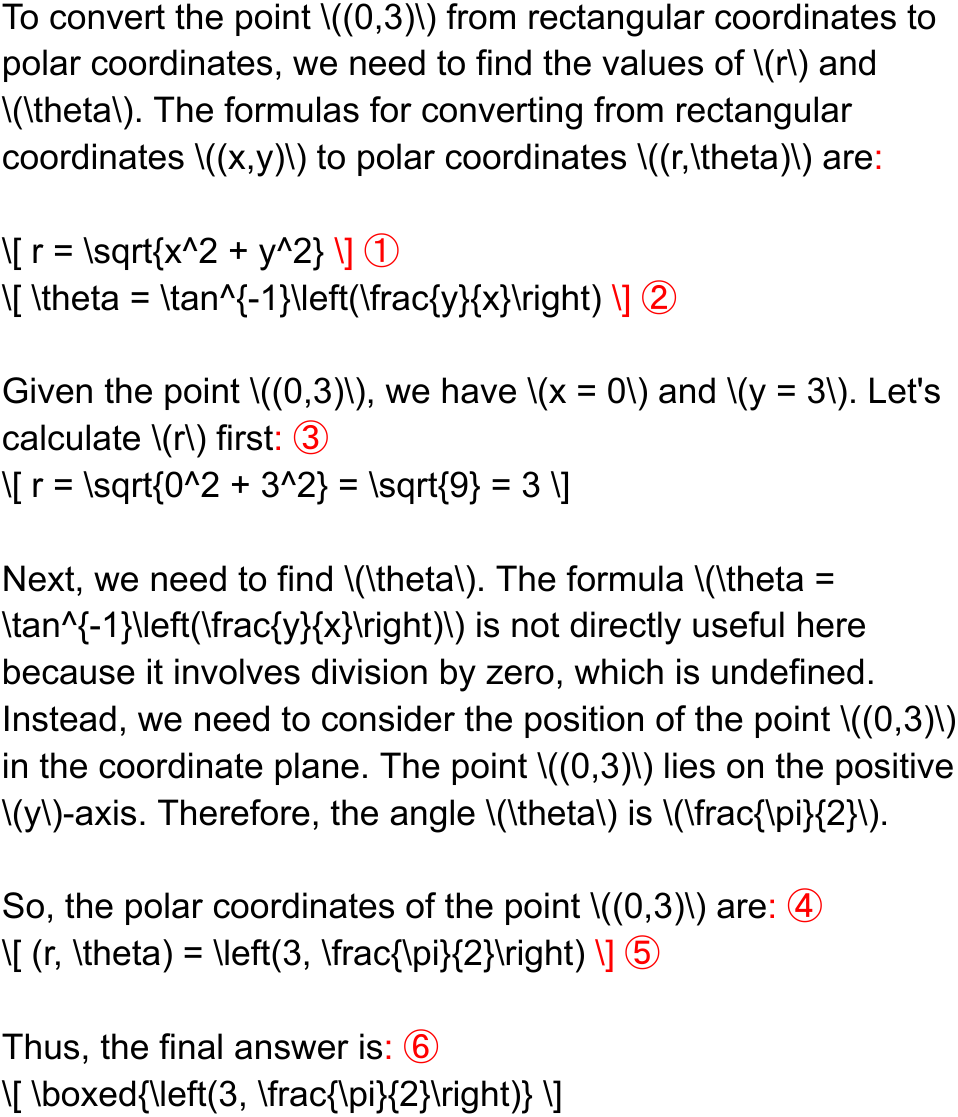}}}
\caption{We input the prefill tokens, ``...Convert the point $(0,3)$ to polar coordinates...'', to Qwen2.5-Math-7B-Instruct and obtain the corresponding decode tokens in the figure. The red tokens represent the milestone tokens or bright columns in Figure~\ref{fig-algo-waterfall} (a).}
\label{fig-algo-milestone}
\end{center}
\vskip -0.2in
\end{figure}

To break the ``impossible trinity'' of sparsity-based algorithms on reasoning tasks, we analyze their decode stage and discover a new attention pattern (Section~\ref{sec-algo-pattern}), based on which we design a new algorithm \algo\ (Section~\ref{sec-algo-algo}) that achieves $O(L)$ time and memory complexities, with accuracy comparable to Quest.


\subsection{Reasoning Attention Pattern}
\label{sec-algo-pattern}

By analyzing the attention map of reasoning tasks' decode stage, we discover two key characteristics (Figure~\ref{fig-algo-waterfall}). First, we identify \textbf{milestone tokens}, which initially exhibit high attention scores but gradually receive lower scores and never receive high scores again. Analogous to lemmas in mathematical proofs, milestone tokens emerge, are utilized, and then fade away. These tokens, visible as bright columns (on the attention map) that slowly diminish (Figure~\ref{fig-algo-waterfall} (a)), must be carefully managed to prevent significant accuracy loss (Figure~\ref{fig-eval-e2e}). Second, we identify \textbf{phoenix tokens}, which receive low attention scores for a period long enough to be evicted from the cache but later regain importance. These tokens typically appear in short prefill prompts, such as user queries (Figure~\ref{fig-algo-waterfall} (b)). Quest~\cite{tang2024quest} retains the entire KV cache to avoid losing phoenix tokens, resulting in its $O(N)$ memory complexity.

We offer a possible explanation for the milestone pattern in reasoning tasks. First, the emergence of milestone tokens is analogous to lemmas in mathematical proofs or subconclusions in thinking steps. Once an LLM generates milestone tokens, subsequent tokens primarily attend to the milestone tokens rather than the preceding tokens arriving at the milestone tokens. Second, the fading attention score of a milestone token mirrors the progression in mathematical reasoning. As reasoning advances from lower-level lemmas to higher-level ones, subsequent steps rely on the new lemmas rather than revisiting the older ones.

To illustrate the preceding explanation, consider one example\footnote{Examples abound during the investigation of reasoning tasks, not limited to this one, and not limited to those extra examples in the appendix.} in Figure~\ref{fig-algo-milestone}. First, tokens \ding{172}\ding{173}\ding{174} serve as initial lemmas, which are crucial for subsequent deductions, corresponding to \ding{172}\ding{173}\ding{174} columns in Figure~\ref{fig-algo-waterfall} (a). Second, tokens \ding{175}\ding{176} serve as a new lemma, built upon \ding{172}\ding{173}\ding{174}, while at the same time, tokens \ding{172}\ding{173}\ding{174} fade. Third, the final answer (token \ding{177}) attend to only the latest milestone tokens \ding{175}\ding{176}.

On the other hand, the definition of phoenix tokens depends on the cache budget: a token qualifies as a phoenix token if it is evicted from the cache and later reused. Under this definition, any token---including milestone tokens---can become a phoenix token. For instance, a milestone token that temporarily receives low attention scores (e.g., for four decoding steps) may be evicted under a cache budget of 4, thus becoming a phoenix token. 

Phoenix tokens exhibit distinct behaviors in prefill and decode tokens. First, phoenix tokens rarely appear in decode tokens when using a sufficiently large cache budget. This behavior arises because decode tokens primarily consist of milestone tokens and other low-importance tokens, and milestone tokens are considered permanently irrelevant if they remain unused for an extended period. A sufficiently large cache budget prevents milestone tokens from becoming phoenix tokens. For example, with a cache budget of 512 tokens, phoenix tokens are rarely observed in decode tokens (Figure~\ref{fig-eval-e2e}). In contrast, with a smaller budget of 64 tokens, phoenix tokens still occur, leading to degraded accuracy for \algo. Second, phoenix tokens frequently occur in prefill tokens regardless of the cache budget. This behavior arises because LLMs typically refer back to the front user queries in the final conclusions of reasoning. To prevent the loss of critical query information, we retain KV vectors of prefill tokens and apply \algo\ (Section~\ref{sec-algo-algo}) exclusively to decode tokens.

\subsection{Design of \algo}
\label{sec-algo-algo}

Based on the preceding observations, we propose a new \algo\ algorithm that addresses the ``impossible trinity'' and consists of two main ideas. First, we identify milestone tokens and retain their KV vectors using a Least-Recently-Used (LRU) caching strategy. During each decoding step, tokens that receive attention scores above the median are considered \textbf{used} and are assigned the latest timestamp. Milestone tokens typically continue to receive the latest timestamps until they become permanently irrelevant. When the KV cache is full, \algo\ evicts KV vectors of the tokens with the oldest timestamp. Second, we retain the KV vectors of all prefill tokens without eviction. Since prefill tokens are typically short and phoenix tokens almost always appear within them in reasoning tasks, retaining these tokens' KV vectors ensures that critical information is not lost in the decode stage.

\begin{algorithm}[t]
\caption{RaaS Algorithm}\label{algo-raas}
\begin{algorithmic}[1]
\STATE \textbf{Input:} A sequence X of prefill tokens, a model M, a KV manager kvm with $r=0.5$
\STATE \textbf{Output:} A sequence Y of prefill tokens plus decode tokens

\STATE Y = X
\STATE y = M.forward(Y, ``prefill'')
\STATE \textbf{while} y $\neq$ eos and len(Y) < M.context\_len()
    \STATE\hspace{0.5cm}Y.append(y)
    \STATE\hspace{0.5cm}y = M.forward(Y, ``decode'')
\STATE \textbf{Return} Y

\STATE 

\STATE \textbf{Function} M.forward(X, stage)
\STATE ...
\STATE K\_old, V\_old = kvm.load\_cache(X)
\STATE Generate Q, K\_new, V\_new for X's new tokens
\vspace{-0.5cm}
\STATE K = K\_old::K\_new, V = V\_old::V\_new
\STATE score = attention(Q, K, V)
\STATE kvm.store\_cache(K, V, score, stage)
\STATE ...
\STATE \textbf{End Function}

\STATE

\STATE \textbf{Function} kvm.store\_cache(K, V, score, stage)
\STATE\textbf{if} stage == ``prefill''
    \STATE\hspace{0.5cm}Store K, V and mark them as non-evictable
    \vspace{-0.5cm}
\STATE\textbf{else if} stage == ``decode''
    \STATE\hspace{0.5cm}Store K\_new and V\_new.
    \STATE\hspace{0.5cm}New timestamps to KVs with top-r score
    \STATE\hspace{0.5cm}\textbf{if} kvm.is\_full()
        \STATE\hspace{0.5cm}\hspace{0.5cm}Evict KVs with the oldest timestamps
\STATE \textbf{End Function}

\end{algorithmic}
\end{algorithm}

Algorithm~\ref{algo-raas} presents the detailed procedure of \algo. (1) Given a sequence X of prefill tokens, the model M first performs the prefill stage, and then proceeds with the decode stage until a stopping criterion is met---such as the generation of an End-Of-Sequence (EOS) token or reaching the model's maximum sequence length. (2) During each forward pass---whether in the prefill or decode stage---M loads the old KVs with a KV manager kvm, computes new KVs, concatenates them (denoted by ``::''), calculates attention scores, and stores the new KVs. (3) The kvm's store\_cache implements the core logic of \algo, as described in the preceding paragraph. It retains a proportion $r$ of tokens with the highest attention scores. Specifically, when $r=0.5$, tokens with attention scores above the median are considered \textbf{used} and receive the latest timestamp.

The choice of $r$ controls the distribution of tokens' timestamps. If $r$ is large, too many tokens will receive the latest timestamp, preventing effective differentiation of milestone tokens. Conversely, if $r$ is small, most tokens are deemed irrelevant, potentially leading to the loss of milestone tokens. To address this dilemma, we propose to assign the latest timestamp to 50\% ($r=0.5$) tokens with the highest attention scores in each decoding step, yielding good results (Figure~\ref{fig-eval-alpha}).

\subsection{Page-Based \algo}

Directly applying the version of \algo\ in Section~\ref{sec-algo-algo} faces two challenges. First, managing KV cache at the token level is inefficient, as small fragmentation in the cache complicates memory management and hinders efficient GPU computation. Second, \algo\ requires the attention scores of all tokens to update timestamps, but retrieving these scores is incompatible with optimized attention kernels such as FlashAttention~\cite{dao2022flashattention, dao2024flashattention2}. As with H2O, bypassing fast kernels in favor of \algo\ could result in degraded performance.

To address these challenges, we propose a page-based version of \algo\footnote{From now on, whenever we use \algo, we refer to page-based \algo.}. First, we introduce a page-based caching system with a fixed page size of $page\_size = 16$ as in vLLM~\cite{woosuk2023vllm}. The timestamp management, as well as cache retention and eviction, is handled at the page level as in most of modern inference engines~\cite{woosuk2023vllm, lianmin2024sglang}. Second, before using optimized attention kernels, we add a lightweight step to retrieve a representative attention score for each page to update its timestamp, similar to Quest. We select a representative Key (K) for each page, and the Query (Q) of the new decoding token attends to these representative keys to compute a single attention score per page. Based on these attention scores, we update the timestamp for each page and make eviction decisions at the page level. Various strategies exist for selecting a representative K, such as those used in Quest~\cite{tang2024quest} and ArkVale~\cite{chenarkvale}. For fairness, we adopt the same representative selection strategy as in Quest.

\begin{figure*}[t]

\begin{center}
\centerline{\includegraphics[width=\textwidth]{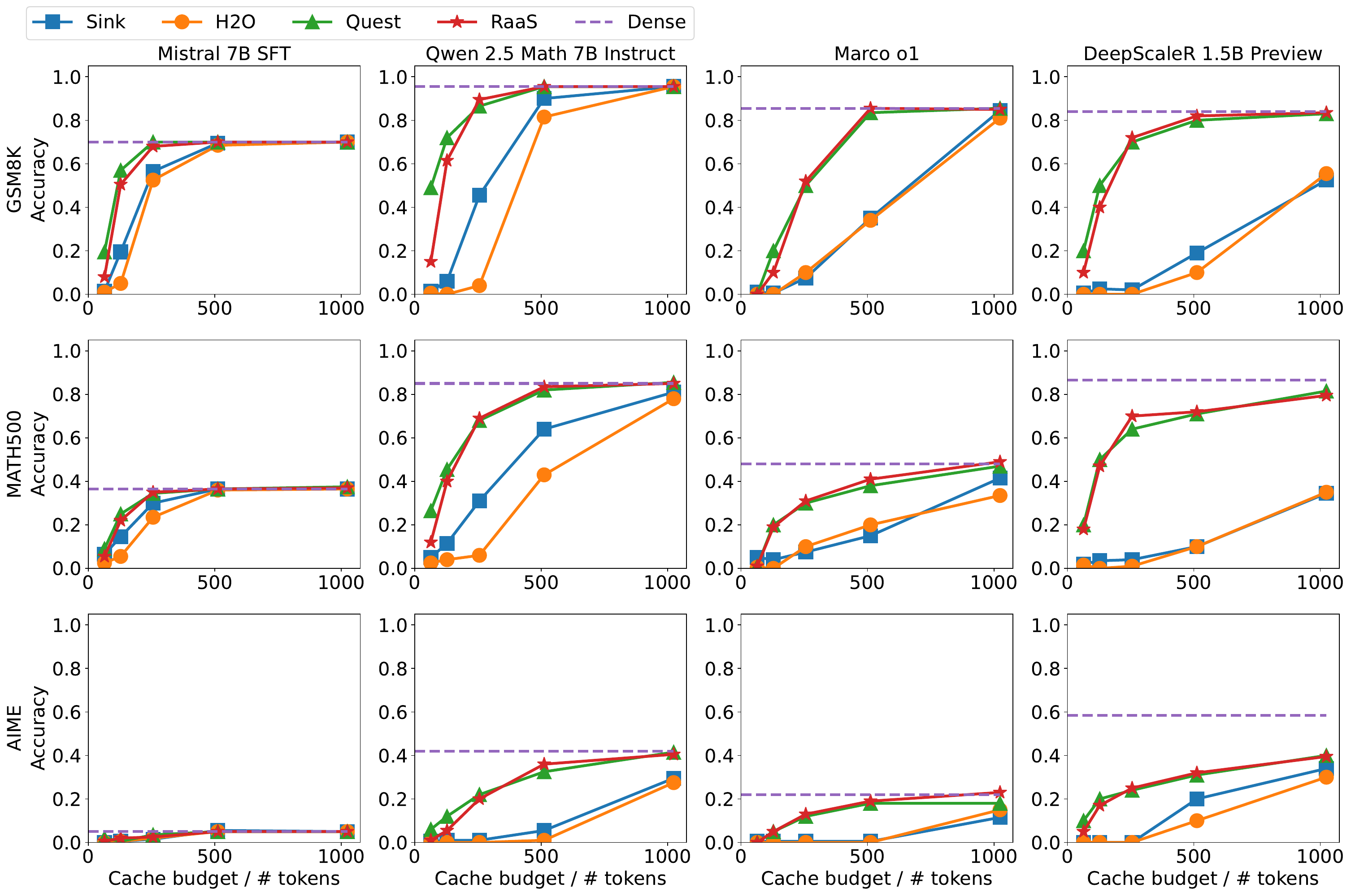}}
\caption{Accuracy vs. cache budget for five algorithms (legends) across three datasets (rows) and four models (columns). The y-axis shows the proportion of correctly solved problems among 200 test cases, while the x-axis represents varying cache budgets: 64, 128, 256, 512, and 1024.}
\label{fig-eval-e2e}
\end{center}
\vskip -0.2in
\end{figure*}
\section{Evaluation}
\label{sec-eval}

We begin by describing the experimental setup, including implementation details, datasets, models, evaluation metrics,
and software/hardware environment. We then present key evaluation results.

\subsection{Experiment Setup}

\textbf{Implementation.} We implement \algo\ based on Hugging Face~\cite{hf} and Quest~\cite{tang2024quest} with 2k lines of Python code. We port Quest from their public repository\footnote{\url{https://github.com/mit-han-lab/Quest}. Accessed on Oct 2024.}.

\textbf{Datasets.} We take the first 200 test cases from each of the following three open-source datasets for our benchmarks: GMS8K~\cite{gms8k}, MATH500~\cite{math500}, and AIME~\cite{aime}, to test the reasoning ability of language models. First, \textit{GMS8k}~\cite{gms8k} contains 8.5k high-quality, linguistically diverse grade-school math problems. These human-written problems need solutions that involve multi-step reasoning and a series of basic arithmetic operations. Second, \textit{MATH500}~\cite{math500} contains 500 challenging problems sourced from high school math competitions with five distinct levels based on the Art of Problem Solving (AoPS) framework, ranging from level 1 to level 5. Third, \textit{AIME}~\cite{aime} is a math problem dataset collected from the American Invitational Mathematics Examination (AIME) competition from 1983 to 2024, designed to challenge the most exceptional high school math students in the United States. These problems cover various fields, such as algebra, geometry, and number theory.

\textbf{Models.} We evaluate our algorithm using four popular models: Marco-o1~\cite{zhao2024marco}, Qwen2.5-Math-7B-Instruct~\cite{wang2024openr}, Mistral-Math-7B~\cite{wang2024openr}, and DeepScaleR-1.5B\footnote{https://pretty-radio-b75.notion.site/DeepScaleR-Surpassing-O1-Preview-with-a-1-5B-Model-by-Scaling-RL-19681902c1468005bed8ca303013a4e2}. They are four of the most powerful open-source LLMs with long-reasoning capabilities.

\textbf{Metrics.} We use two metrics to evaluate performance and model accuracy. First, \textit{Job Completion Time (JCT)} is the time from when users send a request (prompt) to LLMs to when users receive a complete response. A smaller \textit{JCT} indicates a faster algorithm. Second, \textit{Accuracy}~\cite{wang2024openr} measures the mathematical equivalence between an LLM’s output and the ground-truth answer. For each test case, it is either correct or incorrect, and the overall accuracy is reported as the percentage of correctly solved problems across the entire dataset.

\textbf{Baselines.} We compare \algo' accuracy with Dense, H2O, StreamingLLM, and Quest. We implement H2O and StreamingLLM using the HuggingFace Cache class. We compare \algo's latency and memory consumption with only Dense and Quest because StreamingLLM and H2O achieve too low accuracy to be included. We use Quest's official repository with $page\_size=16$.

\textbf{Environment.} We run experiments on a single NVIDIA A100 server with one A100-80GB GPU available. It has 128-core Intel(R) Xeon(R) Platinum 8358P CPU@2.60GHz with two hyperthreading and 1TB DRAM. We use Ubuntu 20.04 with Linux kernel 5.16.7 and CUDA 12.6. Unless stated otherwise, we set $r=0.5$ and $page\_size=16$.

\subsection{Accuracy and Cache Budget Trade-off}

We evaluate five algorithms across three datasets and four models, yielding three key insights from the experimental results (Figure~\ref{fig-eval-e2e}). First, H2O and Sink exhibit poor accuracy under fixed cache budgets compared to others. Sink indiscriminately discards important tokens, including milestone tokens. H2O, on the other hand, overemphasizes accumulated historical attention scores, leading it to retain outdated milestone tokens for too long while discarding newer, relevant ones. Second, Quest and \algo\ achieve the best accuracy. Quest retains KV vectors of all tokens while \algo\ optimizes memory usage by carefully handling milestone tokens with $O(L)$ memory complexity (Figure~\ref{fig-eval-time-memory}). Across these datasets, a cache budget of 1024 tokens is generally sufficient to match Dense’s accuracy. Third, when the cache budget is small, \algo\ underperforms because \algo\ retains all prefill tokens, and with a limited cache budget, most of the budget is allocated to prefill tokens, causing almost all decoding tokens to be discarded, negatively impacting accuracy. For small cache budgets or long-prefill scenarios, we recommend using Quest for prefill tokens and \algo\ for decode tokens.

\subsection{Latency/Memory vs. Decoding Length}

\begin{figure}[t]

\begin{center}
\centerline{\includegraphics[width=\columnwidth]{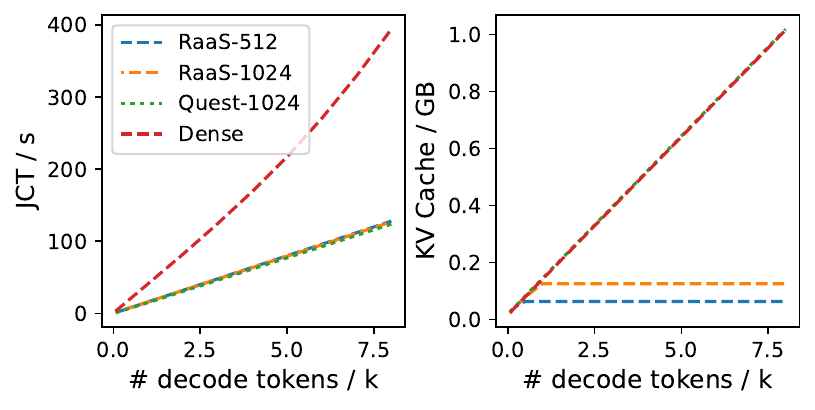}}
\caption{Latency and memory consumptions of Dense, Quest, and \algo\ running on Mistral-Math-7B, using workloads with a fixed prefill length (128 tokens), varying decode lengths (from 0 to 8k tokens) and a batch size of 4. We use dashed lines to improve the visibility of overlapping lines.}
\label{fig-eval-time-memory}
\end{center}
\vskip -0.2in
\end{figure}

We evaluate the Dense, Quest, and \algo\ in terms of their time and memory complexities, yielding two key observations from the experimental results (Figure~\ref{fig-eval-time-memory}). First, as the number of decode tokens increases, Dense's JCT grows quadratically, while both \algo\ and Quest exhibit linear latency growth. The reason is that Dense has $O(N^2)$ time complexity, whereas \algo\ and Quest have $O(NL)$ time complexity, reducing each decoding step from $O(N)$ to $O(L)$. Second, as the number of decode tokens increases, the memory consumption of Dense and Quest grows linearly, while \algo\ initially increases linearly but plateaus once the number of decode tokens exceeds its cache budget. The reason is that Dense and Quest have $O(N)$ memory complexity, whereas \algo\ achieves $O(L)$ memory complexity. With a smaller memory footprint, inference engines using \algo\ are likely to achieve significantly higher throughput.

\subsection{Micro-Benchmarks}

\begin{figure}[t]

\begin{center}
\centerline{\includegraphics[width=\linewidth]{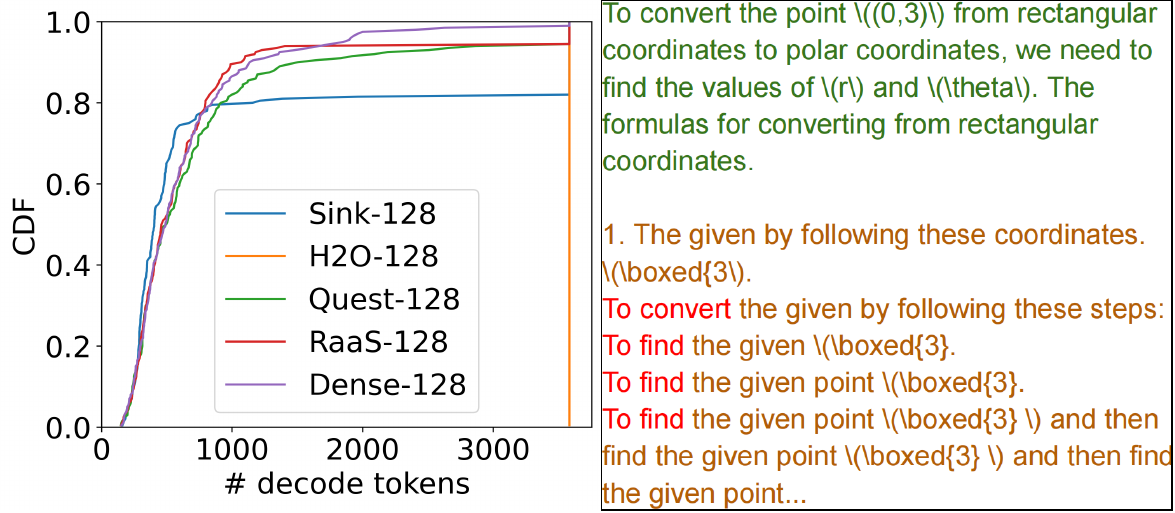}}
\caption{Decode-length distribution of five algorithms using Qwen2.5-Math-7B-Instruct with 4k context length, on MATH500. For example, H2O-128 (128-token cache) always generates to the 4k length limit without solving the problem. On the right, we show a decoding example of H2O-128.}
\label{fig-eval-num-decode}
\end{center}
\vskip -0.2in
\end{figure}

\textbf{The impact of discarding milestone tokens.} Figure~\ref{fig-eval-num-decode} shows that discarding milestone tokens, as in H2O-128 and Sink-128, increases the decode lengths. Sometimes, the decode length increases without solving the problem. Analysis of the outputs reveals that while the model initially reasons correctly for the first few tokens (e.g., green tokens in Figure~\ref{fig-eval-num-decode}), it loses track (orange tokens) of the reasoning process when milestone tokens are discarded, leading to repeated attempts at re-reasoning (red tokens), which ultimately results in the model getting stuck indefinitely.

\begin{figure}[t]

\begin{center}
\centerline{\includegraphics[width=\linewidth]{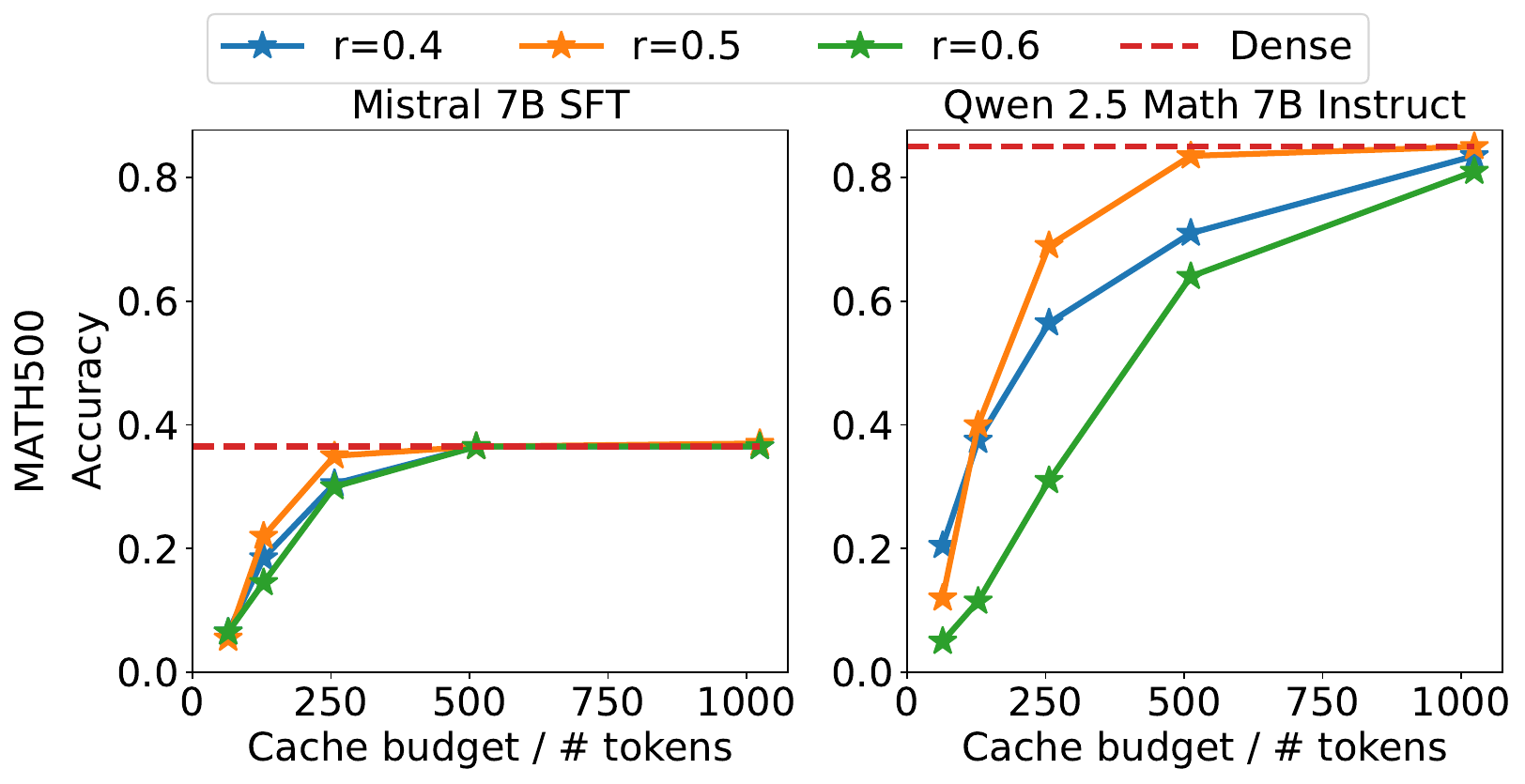}}
\caption{Accuracy of \algo\ with different cache budgets and $r$s.}
\label{fig-eval-alpha}
\end{center}
\vskip -0.2in
\end{figure}
\textbf{The impact of $r$.} The choice of $r$ affects the distribution of tokens' timestamps, with $r=0.5$ generally yielding optimal results, as shown in Figure~\ref{fig-eval-alpha}. First, when $r$ is small, too many tokens are assigned the latest timestamp, preventing effective differentiation of milestone tokens. Second, when $r$ is big, most tokens are deemed irrelevant, potentially leading to the loss of milestone tokens.

\section{Related Work}
\label{sec-related}

Many approaches have been proposed to reduce the time and memory complexities of long-sequence inference; these approaches can be categorized into two types: one that modifies the model architecture and the other that is more plug-and-play.

\subsection{Model Architecture} 

Two types of approaches have emerged for altering model architecture. First, some approaches modify the inner workings of the Transformer while retaining its overall structure. For example, Multi-Query Attention (MQA)~\cite{shazeer2019fast} and Group-Query Attention (GQA)~\cite{ainslie2023gqa} reduce the number of KV heads, achieving similar accuracy to full-head configurations. Second, some approaches change the Transformer architecture significantly in favor of alternative paradigms. For example, RWKV~\cite{peng2023rwkvreinventingrnnstransformer}, RetNet~\cite{sun2023retentivenetworksuccessortransformer}, and Mamba~\cite{gu2024mambalineartimesequencemodeling} adopt RNN-based models, offering lower computational costs but typically underperform compared to Transformer-based models. 

\subsection{KV Compression}

Two primary types of KV compression have emerged: KV quantization and KV pruning. First, KV quantization approaches~\cite{xiao2024smoothquantaccurateefficientposttraining, yao2022zeroquantefficientaffordableposttraining, dettmers2022gpt3,kivi} map higher precision KVs into lower ones, trading accuracy for savings in time and memory. Second, KV pruning approaches focus on leveraging attention sparsity~\cite{zhang2023h2o, ge2023model, jiang2024minference, cai2024pyramidkv, fu2024lazyllm, xiao2024duoattention}, which states that only a few tokens are crucial during LLM inference. Thus, evicting less important tokens from the KV cache is a key strategy for reducing time and memory. For example, StreamingLLM~\cite{xiao2023sink} and LM-Infinite~\cite{han2024lm} evict fixed-position tokens, retaining only the initial and recent window tokens. H2O~\cite{zhang2023h2o}, SnapKV~\cite{li2024snapkv}, ScissorHands~\cite{liu2024scissorhands}, and TOVA~\cite{oren2024transformers} keep the recent tokens and the top-k important tokens based on the attention score calculated within a local window. More recent work, such as Quest~\cite{tang2024quest} and ArkVale~\cite{chenarkvale}, manages the KV cache at the page level, selecting the top-k important pages during each generation step to reduce time complexity.

Our work presents a new trial of applying KV pruning in reasoning tasks, which are characterized by a new milestone attention pattern. For the first time, we achieve real $O(L)$ time and memory complexities with high accuracy.

\section{Conclusion}
\label{sec-conclusion}

In this paper, we have identified a new milestone attention pattern observed in the decode stage of reasoning tasks. Leveraging this pattern, we have proposed a sparsity-based algorithm \algo\ that achieves high accuracy while maintaining $O(L)$ time and $O(L)$ memory complexities. Our experiments, conducted across three datasets and four reasoning-enabled models, demonstrate that \algo\ delivers comparable accuracy and latency to the state-of-the-art Quest, but with constant memory consumption. The key to \algo' success lies in the handling of milestone tokens, which represent intermediate conclusions leading to the final output.

\balance
\section*{Limitations}
\label{sec-limitations}

Our work in this paper has the following major limitations.

\textbf{Lack of comprehensive attention map statistics.} Attention patterns vary across layers, heads, and model architectures. Manual inspection, without rigorous statistical analysis, is insufficient for drawing general conclusions about emerging attention patterns. Our study remains small-scale due to the absence of an automated tool for analyzing attention patterns across datasets, models, and layers. Such a tool would take an input defining an attention pattern, a range of datasets and models, and output the statistics of the specified attention pattern across each model's layers and heads. Although we cannot address this limitation for now, we plan to develop such a tool in the future, and this tool would benefit the entire research community.

\textbf{Limited applicability of \algo.} \algo\ is specifically designed for reasoning tasks where the number of prefill tokens (e.g., a mathematical query) is small but the number of decode tokens (e.g., a chain of reasoning followed by a final answer) is large. Given that \algo\ retains KV vectors of all prefill tokens, it allocates the entire cache budget to them and discards nearly all decode tokens in long-prefill scenarios. Therefore, in these scenarios, we recommend using the combination of Quest (on only prefill tokens) and \algo\ (on only decode tokens). 

\textbf{Evaluation on a limited set of datasets and models.} Our evaluation covers only four models and three datasets. As such, the results may not generalize beyond these specific configurations. Although models with longer context lengths (e.g., Qwen2.5-Max, DeepSeek-r1) and datasets such as GPQA Diamond and Codeforces exist, exhaustive evaluation across all combinations is computationally prohibitive~\cite{hu2023pcrml}. As reported in prior work~\cite{zhong2024distserve}, decoding a single token can take approximately 30 ms; thus, processing 16k tokens on an A100-80GB GPU requires around 8 minutes. Running 200 test cases would take over a day on a single GPU, making large-scale evaluation infeasible with limited resources. Nonetheless, we believe that the core idea of the milestone pattern and its underlying rationale---the thinking/reasoning process is step-by-step, constructing lemmas along the way--- remains broadly applicable.

\section*{Acknowledgments}

This work was partially supported by National Natural Science Foundation of China under Grant No. 92464301. We would also like to thank the anonymous reviewers for their insightful comments and suggestions, which help improve the quality of this paper.

\newpage
\bibliography{custom}

\begin{thebibliography}{52}
\providecommand{\natexlab}[1]{#1}

\bibitem[{{AIME}()}]{aime}
{AIME}.
\newblock {AIME\_1983\_2024}.
\newblock \url{https://huggingface.co/datasets/di-zhang-fdu/AIME\_1983\_2024}.

\bibitem[{Ainslie et~al.(2023)Ainslie, Lee{-}Thorp, de~Jong, Zemlyanskiy, Lebr{\'{o}}n, and Sanghai}]{ainslie2023gqa}
Joshua Ainslie, James Lee{-}Thorp, Michiel de~Jong, Yury Zemlyanskiy, Federico Lebr{\'{o}}n, and Sumit Sanghai. 2023.
\newblock {GQA}: Training generalized multi-query transformer models from multi-head checkpoints.
\newblock In \emph{Proceedings of the 2023 Conference on Empirical Methods in Natural Language Processing}, pages 4895--4901.

\bibitem[{Bai et~al.(2024)Bai, Lv, Zhang, Lyu, Tang, Huang, Du, Liu, Zeng, Hou, Dong, Tang, and Li}]{yushi2024longbench}
Yushi Bai, Xin Lv, Jiajie Zhang, Hongchang Lyu, Jiankai Tang, Zhidian Huang, Zhengxiao Du, Xiao Liu, Aohan Zeng, Lei Hou, Yuxiao Dong, Jie Tang, and Juanzi Li. 2024.
\newblock {LongBench}: {A} bilingual, multitask benchmark for long context understanding.
\newblock In \emph{Proceedings of the Sixty-Second Annual Meeting of the Association for Computational Linguistics}, pages 3119--3137.

\bibitem[{Cai et~al.(2024)Cai, Zhang, Gao, Liu, Liu, Lu, Xiong, Dong, Chang, Hu, and Xiao}]{cai2024pyramidkv}
Zefan Cai, Yichi Zhang, Bofei Gao, Yuliang Liu, Tianyu Liu, Keming Lu, Wayne Xiong, Yue Dong, Baobao Chang, Junjie Hu, and Wen Xiao. 2024.
\newblock {PyramidKV}: Dynamic {KV} cache compression based on pyramidal information funneling.
\newblock \emph{CoRR}.

\bibitem[{Chen et~al.(2024)Chen, Wang, Cao, Wu, Zheng, Li, Wei, Yan, Li, and Liang}]{chenarkvale}
Renze Chen, Zhuofeng Wang, Beiquan Cao, Tong Wu, Size Zheng, Xiuhong Li, Xuechao Wei, Shengen Yan, Meng Li, and Yun Liang. 2024.
\newblock {ArkVale}: Efficient generative {LLM} inference with recallable key-value eviction.
\newblock In \emph{Proceedings of the Thirty-Eighth Annual Conference on Neural Information Processing Systems}, pages 113134--113155.

\bibitem[{Cobbe et~al.(2021)Cobbe, Kosaraju, Bavarian, Chen, Jun, Kaiser, Plappert, Tworek, Hilton, Nakano, Hesse, and Schulman}]{gms8k}
Karl Cobbe, Vineet Kosaraju, Mohammad Bavarian, Mark Chen, Heewoo Jun, Lukasz Kaiser, Matthias Plappert, Jerry Tworek, Jacob Hilton, Reiichiro Nakano, Christopher Hesse, and John Schulman. 2021.
\newblock Training verifiers to solve math word problems.
\newblock \emph{CoRR}.

\bibitem[{Dai et~al.(2024)Dai, Deng, Zhao, Xu, Gao, Chen, Li, Zeng, Yu, Wu, Xie, Li, Huang, Luo, Ruan, Sui, and Liang}]{damai2024deepseek}
Damai Dai, Chengqi Deng, Chenggang Zhao, R.~X. Xu, Huazuo Gao, Deli Chen, Jiashi Li, Wangding Zeng, Xingkai Yu, Y.~Wu, Zhenda Xie, Y.~K. Li, Panpan Huang, Fuli Luo, Chong Ruan, Zhifang Sui, and Wenfeng Liang. 2024.
\newblock {DeepSeekMoE}: Towards ultimate expert specialization in mixture-of-experts language models.
\newblock In \emph{Proceedings of the Sixty-Second Annual Meeting of the Association for Computational Linguistics}, pages 1280--1297.

\bibitem[{Dao(2024)}]{dao2024flashattention2}
Tri Dao. 2024.
\newblock {FlashAttention-2}: Faster attention with better parallelism and work partitioning.
\newblock In \emph{Proceedings of the Twelfth International Conference on Learning Representations}.

\bibitem[{Dao et~al.(2022)Dao, Fu, Ermon, Rudra, and R{\'{e}}}]{dao2022flashattention}
Tri Dao, Daniel~Y. Fu, Stefano Ermon, Atri Rudra, and Christopher R{\'{e}}. 2022.
\newblock {FlashAttention}: Fast and memory-efficient exact attention with {IO}-awareness.
\newblock In \emph{Proceedings of the Thirty-Sixth Annual Conference on Neural Information Processing Systems}, pages 16344--16359.

\bibitem[{Dettmers et~al.(2022)Dettmers, Lewis, Belkada, and Zettlemoyer}]{dettmers2022gpt3}
Tim Dettmers, Mike Lewis, Younes Belkada, and Luke Zettlemoyer. 2022.
\newblock {GPT3.int8()}: 8-bit matrix multiplication for transformers at scale.
\newblock In \emph{Proceedings of the Thirty-Sixth Annual Conference on Neural Information Processing Systems}, pages 30318--30332.

\bibitem[{Fu et~al.(2024)Fu, Cho, Merth, Mehta, Rastegari, and Najibi}]{fu2024lazyllm}
Qichen Fu, Minsik Cho, Thomas Merth, Sachin Mehta, Mohammad Rastegari, and Mahyar Najibi. 2024.
\newblock {LazyLLM}: Dynamic token pruning for efficient long context {LLM} inference.
\newblock \emph{CoRR}.

\bibitem[{Gao et~al.(2023)Gao, Xiong, Gao, Jia, Pan, Bi, Dai, Sun, Guo, Wang, and Wang}]{gao2023retrieval}
Yunfan Gao, Yun Xiong, Xinyu Gao, Kangxiang Jia, Jinliu Pan, Yuxi Bi, Yi~Dai, Jiawei Sun, Qianyu Guo, Meng Wang, and Haofen Wang. 2023.
\newblock Retrieval-augmented generation for large language models: {A} survey.
\newblock \emph{CoRR}.

\bibitem[{Ge et~al.(2024)Ge, Zhang, Liu, Zhang, Han, and Gao}]{ge2023model}
Suyu Ge, Yunan Zhang, Liyuan Liu, Minjia Zhang, Jiawei Han, and Jianfeng Gao. 2024.
\newblock Model tells you what to discard: Adaptive {KV} cache compression for {LLMs}.
\newblock In \emph{Proceedings of the Twelfth International Conference on Learning Representations}.

\bibitem[{Gu and Dao(2023)}]{gu2024mambalineartimesequencemodeling}
Albert Gu and Tri Dao. 2023.
\newblock {Mamba}: Linear-time sequence modeling with selective state spaces.
\newblock \emph{CoRR}.

\bibitem[{Han et~al.(2024)Han, Wang, Peng, Xiong, Chen, Ji, and Wang}]{han2024lm}
Chi Han, Qifan Wang, Hao Peng, Wenhan Xiong, Yu~Chen, Heng Ji, and Sinong Wang. 2024.
\newblock {LM-Infinite}: Zero-shot extreme length generalization for large language models.
\newblock In \emph{Proceedings of the 2024 Conference of the North American Chapter of the Association for Computational Linguistics: Human Language Technologies}, pages 3991--4008.

\bibitem[{Hendrycks et~al.(2021)Hendrycks, Burns, Kadavath, Arora, Basart, Tang, Song, and Steinhardt}]{math500}
Dan Hendrycks, Collin Burns, Saurav Kadavath, Akul Arora, Steven Basart, Eric Tang, Dawn Song, and Jacob Steinhardt. 2021.
\newblock Measuring mathematical problem solving with the {MATH} dataset.
\newblock In \emph{Proceedings of the Neural Information Processing Systems Track on Datasets and Benchmarks}.

\bibitem[{Hu et~al.(2024)Hu, Huang, Hu, Xu, Chen, Xie, Wang, Wang, Bao, Sun, and Shan}]{cunchen2024memserve}
Cunchen Hu, Heyang Huang, Junhao Hu, Jiang Xu, Xusheng Chen, Tao Xie, Chenxi Wang, Sa~Wang, Yungang Bao, Ninghui Sun, and Yizhou Shan. 2024.
\newblock {MemServe}: Context caching for disaggregated {LLM} serving with elastic memory pool.
\newblock \emph{CoRR}.

\bibitem[{Hu et~al.(2025{\natexlab{a}})Hu, Huang, Wang, Wang, Hu, Zhang, Feng, Chen, Shan, and Xie}]{hu2025epic}
Junhao Hu, Wenrui Huang, Haoyi Wang, Weidong Wang, Tiancheng Hu, Qin Zhang, Hao Feng, Xusheng Chen, Yizhou Shan, and Tao Xie. 2025{\natexlab{a}}.
\newblock {EPIC:} efficient position-independent caching for serving large language models.
\newblock In \emph{Proceedings of the Forty-Second International Conference on Machine Learning}.

\bibitem[{Hu et~al.(2023)Hu, Wang, Huang, Luo, Jin, Deng, and Xie}]{hu2023pcrml}
Junhao Hu, Chaozheng Wang, Hailiang Huang, Huang Luo, Yu~Jin, Yuetang Deng, and Tao Xie. 2023.
\newblock Predicting compilation resources for adaptive build in an industrial setting.
\newblock In \emph{Proceedings of the Thity-Eighth {IEEE/ACM} International Conference on Automated Software Engineering}, pages 1808--1813.

\bibitem[{Hu et~al.(2025{\natexlab{b}})Hu, Xu, Liu, He, Chen, Xu, Liu, Zhang, Wan, Dan, Dong, Ren, Meng, He, Liu, Xie, Lin, Zhang, Yu, Feng, Chen, and Shan}]{hu2025deepflow}
Junhao Hu, Jiang Xu, Zhixia Liu, Yulong He, Yuetao Chen, Hao Xu, Jiang Liu, Baoquan Zhang, Shining Wan, Gengyuan Dan, Zhiyu Dong, Zhihao Ren, Jie Meng, Chao He, Changhong Liu, Tao Xie, Dayun Lin, Qin Zhang, Yue Yu, Hao Feng, Xusheng Chen, and Yizhou Shan. 2025{\natexlab{b}}.
\newblock {DEEPSERVE}: Serverless large language model serving at scale.
\newblock In \emph{Proceedings of the 2025 {USENIX} Annual Technical Conference}.

\bibitem[{{Hugging Face}()}]{hf}
{Hugging Face}.
\newblock {Hugging Face}.
\newblock \url{https://huggingface.co}.

\bibitem[{Jeong et~al.(2024)Jeong, Baek, Cho, Hwang, and Park}]{jeong2024adaptive}
Soyeong Jeong, Jinheon Baek, Sukmin Cho, Sung~Ju Hwang, and Jong Park. 2024.
\newblock Adaptive-{RAG}: Learning to adapt retrieval-augmented large language models through question complexity.
\newblock In \emph{Proceedings of the 2024 Conference of the North American Chapter of the Association for Computational Linguistics: Human Language Technologies}, pages 7036--7050.

\bibitem[{Jiang et~al.(2024)Jiang, Li, Zhang, Wu, Luo, Ahn, Han, Abdi, Li, Lin, Yang, and Qiu}]{jiang2024minference}
Huiqiang Jiang, Yucheng Li, Chengruidong Zhang, Qianhui Wu, Xufang Luo, Surin Ahn, Zhenhua Han, Amir Abdi, Dongsheng Li, Chin{-}Yew Lin, Yuqing Yang, and Lili Qiu. 2024.
\newblock {MInference} 1.0: Accelerating pre-filling for long-context {LLMs} via dynamic sparse attention.
\newblock In \emph{Proceedings of the Thirty-Eighth Annual Conference on Neural Information Processing Systems}, pages 52481--52515.

\bibitem[{Jin et~al.(2024)Jin, Zhang, Jiang, Liu, Liu, Liu, and Jin}]{jin2024ragcache}
Chao Jin, Zili Zhang, Xuanlin Jiang, Fangyue Liu, Xin Liu, Xuanzhe Liu, and Xin Jin. 2024.
\newblock {RAGCache}: Efficient knowledge caching for retrieval-augmented generation.
\newblock \emph{CoRR}.

\bibitem[{Kwon et~al.(2023)Kwon, Li, Zhuang, Sheng, Zheng, Yu, Gonzalez, Zhang, and Stoica}]{woosuk2023vllm}
Woosuk Kwon, Zhuohan Li, Siyuan Zhuang, Ying Sheng, Lianmin Zheng, Cody~Hao Yu, Joseph Gonzalez, Hao Zhang, and Ion Stoica. 2023.
\newblock Efficient memory management for large language model serving with {PagedAttention}.
\newblock In \emph{Proceedings of the Twenty-Ninth Symposium on Operating Systems Principles}, pages 611--626.

\bibitem[{Li et~al.(2022)Li, Su, Cai, Wang, and Liu}]{li2022survey}
Huayang Li, Yixuan Su, Deng Cai, Yan Wang, and Lemao Liu. 2022.
\newblock A survey on retrieval-augmented text generation.
\newblock \emph{CoRR}.

\bibitem[{Li et~al.(2024)Li, Huang, Yang, Venkitesh, Locatelli, Ye, Cai, Lewis, and Chen}]{li2024snapkv}
Yuhong Li, Yingbing Huang, Bowen Yang, Bharat Venkitesh, Acyr Locatelli, Hanchen Ye, Tianle Cai, Patrick Lewis, and Deming Chen. 2024.
\newblock {SnapKV}: {LLM} knows what you are looking for before generation.
\newblock In \emph{Proceedings of the Thirty-Eighth Annual Conference on Neural Information Processing Systems}, pages 22947--22970.

\bibitem[{Lightman et~al.(2024)Lightman, Kosaraju, Burda, Edwards, Baker, Lee, Leike, Schulman, Sutskever, and Cobbe}]{lightman2023let}
Hunter Lightman, Vineet Kosaraju, Yuri Burda, Harrison Edwards, Bowen Baker, Teddy Lee, Jan Leike, John Schulman, Ilya Sutskever, and Karl Cobbe. 2024.
\newblock Let's verify step by step.
\newblock In \emph{Proceedings of the Twelfth International Conference on Learning Representations}.

\bibitem[{Liu et~al.(2023)Liu, Desai, Liao, Wang, Xie, Xu, Kyrillidis, and Shrivastava}]{liu2024scissorhands}
Zichang Liu, Aditya Desai, Fangshuo Liao, Weitao Wang, Victor Xie, Zhaozhuo Xu, Anastasios Kyrillidis, and Anshumali Shrivastava. 2023.
\newblock {Scissorhands}: Exploiting the persistence of importance hypothesis for {LLM} {KV} cache compression at test time.
\newblock In \emph{Proceedings of the Thirty-Seventh Annual Conference on Neural Information Processing Systems}, pages 52342--52364.

\bibitem[{Liu et~al.(2024)Liu, Yuan, Jin, Zhong, Xu, Braverman, Chen, and Hu}]{kivi}
Zirui Liu, Jiayi Yuan, Hongye Jin, Shaochen Zhong, Zhaozhuo Xu, Vladimir Braverman, Beidi Chen, and Xia Hu. 2024.
\newblock {KIVI:} {A} tuning-free asymmetric 2bit quantization for {KV} cache.
\newblock In \emph{Proceedings of the Forty-First International Conference on Machine Learning}, pages 32332--32344.

\bibitem[{Mao et~al.(2021)Mao, He, Liu, Shen, Gao, Han, and Chen}]{mao2020generation}
Yuning Mao, Pengcheng He, Xiaodong Liu, Yelong Shen, Jianfeng Gao, Jiawei Han, and Weizhu Chen. 2021.
\newblock Generation-augmented retrieval for open-domain question answering.
\newblock In \emph{Proceedings of the Fifty-Ninth Annual Meeting of the Association for Computational Linguistics and the Eleventh International Joint Conference on Natural Language Processing}, pages 4089--4100.

\bibitem[{{OpenAI}()}]{openai2024o1}
{OpenAI}.
\newblock {OpenAI o1}.
\newblock \url{https://openai.com/o1/}.

\bibitem[{Oren et~al.(2024)Oren, Hassid, Nir, Adi, and Schwartz}]{oren2024transformers}
Matanel Oren, Michael Hassid, Yarden Nir, Yossi Adi, and Roy Schwartz. 2024.
\newblock Transformers are multi-state {RNNs}.
\newblock In \emph{Proceedings of the 2024 Conference on Empirical Methods in Natural Language Processing}, pages 18724--18741.

\bibitem[{Peng et~al.(2023)Peng, Alcaide, Anthony, Albalak, Arcadinho, Biderman, Cao, Cheng, Chung, Derczynski, Du, Grella, GV, He, Hou, Kazienko, Kocon, Kong, Koptyra, Lau, Lin, Mantri, Mom, Saito, Song, Tang, Wind, Wozniak, Zhang, Zhou, Zhu, and Zhu}]{peng2023rwkvreinventingrnnstransformer}
Bo~Peng, Eric Alcaide, Quentin Anthony, Alon Albalak, Samuel Arcadinho, Stella Biderman, Huanqi Cao, Xin Cheng, Michael Chung, Leon Derczynski, Xingjian Du, Matteo Grella, Kranthi~Kiran GV, Xuzheng He, Haowen Hou, Przemyslaw Kazienko, Jan Kocon, Jiaming Kong, Bartlomiej Koptyra, Hayden Lau, Jiaju Lin, Krishna Sri~Ipsit Mantri, Ferdinand Mom, Atsushi Saito, Guangyu Song, Xiangru Tang, Johan~S. Wind, Stanislaw Wozniak, Zhenyuan Zhang, Qinghua Zhou, Jian Zhu, and Rui{-}Jie Zhu. 2023.
\newblock {RWKV}: Reinventing {RNNs} for the transformer era.
\newblock In \emph{Findings of the Association for Computational Linguistics: {EMNLP}}, pages 14048--14077.

\bibitem[{Pope et~al.(2023)Pope, Douglas, Chowdhery, Devlin, Bradbury, Heek, Xiao, Agrawal, and Dean}]{pope2023efficiently}
Reiner Pope, Sholto Douglas, Aakanksha Chowdhery, Jacob Devlin, James Bradbury, Jonathan Heek, Kefan Xiao, Shivani Agrawal, and Jeff Dean. 2023.
\newblock Efficiently scaling transformer inference.
\newblock In \emph{Proceedings of the Sixth Conference on Machine Learning and Systems}, pages 606--624.

\bibitem[{Ram et~al.(2023)Ram, Levine, Dalmedigos, Muhlgay, Shashua, Leyton{-}Brown, and Shoham}]{ram2023context}
Ori Ram, Yoav Levine, Itay Dalmedigos, Dor Muhlgay, Amnon Shashua, Kevin Leyton{-}Brown, and Yoav Shoham. 2023.
\newblock In-context retrieval-augmented language models.
\newblock \emph{Transactions of the Association for Computational Linguistics}.

\bibitem[{Shazeer(2019)}]{shazeer2019fast}
Noam Shazeer. 2019.
\newblock Fast transformer decoding: One write-head is all you need.
\newblock \emph{CoRR}.

\bibitem[{Sun et~al.(2023)Sun, Dong, Huang, Ma, Xia, Xue, Wang, and Wei}]{sun2023retentivenetworksuccessortransformer}
Yutao Sun, Li~Dong, Shaohan Huang, Shuming Ma, Yuqing Xia, Jilong Xue, Jianyong Wang, and Furu Wei. 2023.
\newblock Retentive network: {A} successor to transformer for large language models.
\newblock \emph{CoRR}.

\bibitem[{Tang et~al.(2024)Tang, Zhao, Zhu, Xiao, Kasikci, and Han}]{tang2024quest}
Jiaming Tang, Yilong Zhao, Kan Zhu, Guangxuan Xiao, Baris Kasikci, and Song Han. 2024.
\newblock {QUEST:} query-aware sparsity for efficient long-context {LLM} inference.
\newblock In \emph{Proceedings of the Forty-First International Conference on Machine Learning}, pages 47901--47911.

\bibitem[{Vaswani et~al.(2017)Vaswani, Shazeer, Parmar, Uszkoreit, Jones, Gomez, Kaiser, and Polosukhin}]{vaswani2017attention}
Ashish Vaswani, Noam Shazeer, Niki Parmar, Jakob Uszkoreit, Llion Jones, Aidan~N. Gomez, Lukasz Kaiser, and Illia Polosukhin. 2017.
\newblock Attention is all you need.
\newblock In \emph{Proceedings of the Thirty-First Annual Conference on Neural Information Processing Systems}, pages 5998--6008.

\bibitem[{Wang et~al.(2024)Wang, Fang, Wan, Wen, Zhu, Liu, Gong, Song, Chen, Ni, Yang, Wen, and Zhang}]{wang2024openr}
Jun Wang, Meng Fang, Ziyu Wan, Muning Wen, Jiachen Zhu, Anjie Liu, Ziqin Gong, Yan Song, Lei Chen, Lionel~M. Ni, Linyi Yang, Ying Wen, and Weinan Zhang. 2024.
\newblock {OpenR}: An open source framework for advanced reasoning with large language models.
\newblock \emph{CoRR}.

\bibitem[{Wei et~al.(2022)Wei, Wang, Schuurmans, Bosma, Ichter, Xia, Chi, Le, and Zhou}]{wei2022chain}
Jason Wei, Xuezhi Wang, Dale Schuurmans, Maarten Bosma, Brian Ichter, Fei Xia, Ed~H. Chi, Quoc~V. Le, and Denny Zhou. 2022.
\newblock Chain-of-thought prompting elicits reasoning in large language models.
\newblock In \emph{Proceedings of the Thirty-Sixth Annual Conference on Neural Information Processing Systems}, pages 24824--24837.

\bibitem[{Xiao et~al.(2023)Xiao, Lin, Seznec, Wu, Demouth, and Han}]{xiao2024smoothquantaccurateefficientposttraining}
Guangxuan Xiao, Ji~Lin, Micka{\"{e}}l Seznec, Hao Wu, Julien Demouth, and Song Han. 2023.
\newblock {SmoothQuant}: Accurate and efficient post-training quantization for large language models.
\newblock In \emph{Proceedings of the Fortieth International Conference on Machine Learning}, pages 38087--38099.

\bibitem[{Xiao et~al.(2025)Xiao, Tang, Zuo, Guo, Yang, Tang, Fu, and Han}]{xiao2024duoattention}
Guangxuan Xiao, Jiaming Tang, Jingwei Zuo, Junxian Guo, Shang Yang, Haotian Tang, Yao Fu, and Song Han. 2025.
\newblock {DuoAttention}: Efficient long-context {LLM} inference with retrieval and streaming heads.
\newblock In \emph{Proceedings of the Thirteenth International Conference on Learning Representations}.

\bibitem[{Xiao et~al.(2024)Xiao, Tian, Chen, Han, and Lewis}]{xiao2023sink}
Guangxuan Xiao, Yuandong Tian, Beidi Chen, Song Han, and Mike Lewis. 2024.
\newblock Efficient streaming language models with attention sinks.
\newblock In \emph{Proceedings of the Twelfth International Conference on Learning Representations}.

\bibitem[{Yang et~al.(2024)Yang, Zhang, Hui, Gao, Yu, Li, Liu, Tu, Zhou, Lin, Lu, Xue, Lin, Liu, Ren, and Zhang}]{qwen25mathtechnical}
An~Yang, Beichen Zhang, Binyuan Hui, Bofei Gao, Bowen Yu, Chengpeng Li, Dayiheng Liu, Jianhong Tu, Jingren Zhou, Junyang Lin, Keming Lu, Mingfeng Xue, Runji Lin, Tianyu Liu, Xingzhang Ren, and Zhenru Zhang. 2024.
\newblock {Qwen2.5-Math} technical report: Toward mathematical expert model via self-improvement.
\newblock \emph{CoRR}.

\bibitem[{Yao et~al.(2022)Yao, Aminabadi, Zhang, Wu, Li, and He}]{yao2022zeroquantefficientaffordableposttraining}
Zhewei Yao, Reza~Yazdani Aminabadi, Minjia Zhang, Xiaoxia Wu, Conglong Li, and Yuxiong He. 2022.
\newblock {ZeroQuant}: Efficient and affordable post-training quantization for large-scale transformers.
\newblock In \emph{Proceedings of the Thirty-Sixth Annual Conference on Neural Information Processing Systems}, pages 27168--27183.

\bibitem[{Zhang et~al.(2022)Zhang, Zhang, Du, Du, Pang, Gao, and Lin}]{zhang2025lighttransfer}
Xuan Zhang, Fengzhuo Zhang, Cunxiao Du, Chao Du, Tianyu Pang, Wei Gao, and Min Lin. 2022.
\newblock {LightTransfer:} your long-context {LLM} is secretly a hybrid model with effortless adaptation.
\newblock \emph{CoRR}.

\bibitem[{Zhang et~al.(2023)Zhang, Sheng, Zhou, Chen, Zheng, Cai, Song, Tian, R{\'{e}}, Barrett, Wang, and Chen}]{zhang2023h2o}
Zhenyu Zhang, Ying Sheng, Tianyi Zhou, Tianlong Chen, Lianmin Zheng, Ruisi Cai, Zhao Song, Yuandong Tian, Christopher R{\'{e}}, Clark~W. Barrett, Zhangyang Wang, and Beidi Chen. 2023.
\newblock {H2O:} heavy-hitter oracle for efficient generative inference of large language models.
\newblock In \emph{Proceedings of the Thirty-Seventh Annual Conference on Neural Information Processing Systems}, pages 34661--34710.

\bibitem[{Zhao et~al.(2024)Zhao, Yin, Zeng, Wang, Shi, Lyu, Wang, Luo, and Zhang}]{zhao2024marco}
Yu~Zhao, Huifeng Yin, Bo~Zeng, Hao Wang, Tianqi Shi, Chenyang Lyu, Longyue Wang, Weihua Luo, and Kaifu Zhang. 2024.
\newblock {Marco-o1}: Towards open reasoning models for open-ended solutions.
\newblock \emph{CoRR}.

\bibitem[{Zheng et~al.(2024)Zheng, Yin, Xie, Sun, Huang, Yu, Cao, Kozyrakis, Stoica, Gonzalez, Barrett, and Sheng}]{lianmin2024sglang}
Lianmin Zheng, Liangsheng Yin, Zhiqiang Xie, Chuyue Sun, Jeff Huang, Cody~Hao Yu, Shiyi Cao, Christos Kozyrakis, Ion Stoica, Joseph~E. Gonzalez, Clark~W. Barrett, and Ying Sheng. 2024.
\newblock {SGLang}: Efficient execution of structured language model programs.
\newblock In \emph{Proceedings of the Thirty-Eighth Annual Conference on Neural Information Processing Systems}, pages 62557--62583.

\bibitem[{Zhong et~al.(2024)Zhong, Liu, Chen, Hu, Zhu, Liu, Jin, and Zhang}]{zhong2024distserve}
Yinmin Zhong, Shengyu Liu, Junda Chen, Jianbo Hu, Yibo Zhu, Xuanzhe Liu, Xin Jin, and Hao Zhang. 2024.
\newblock {DistServe}: Disaggregating prefill and decoding for goodput-optimized large language model serving.
\newblock In \emph{Proceedings of the Eighteenth {USENIX} Symposium on Operating Systems Design and Implementation}, pages 193--210.

\end{thebibliography}

\newpage
\appendix

\begin{figure}[th]

\begin{center}
\fbox{\centerline{\includegraphics[width=\columnwidth]{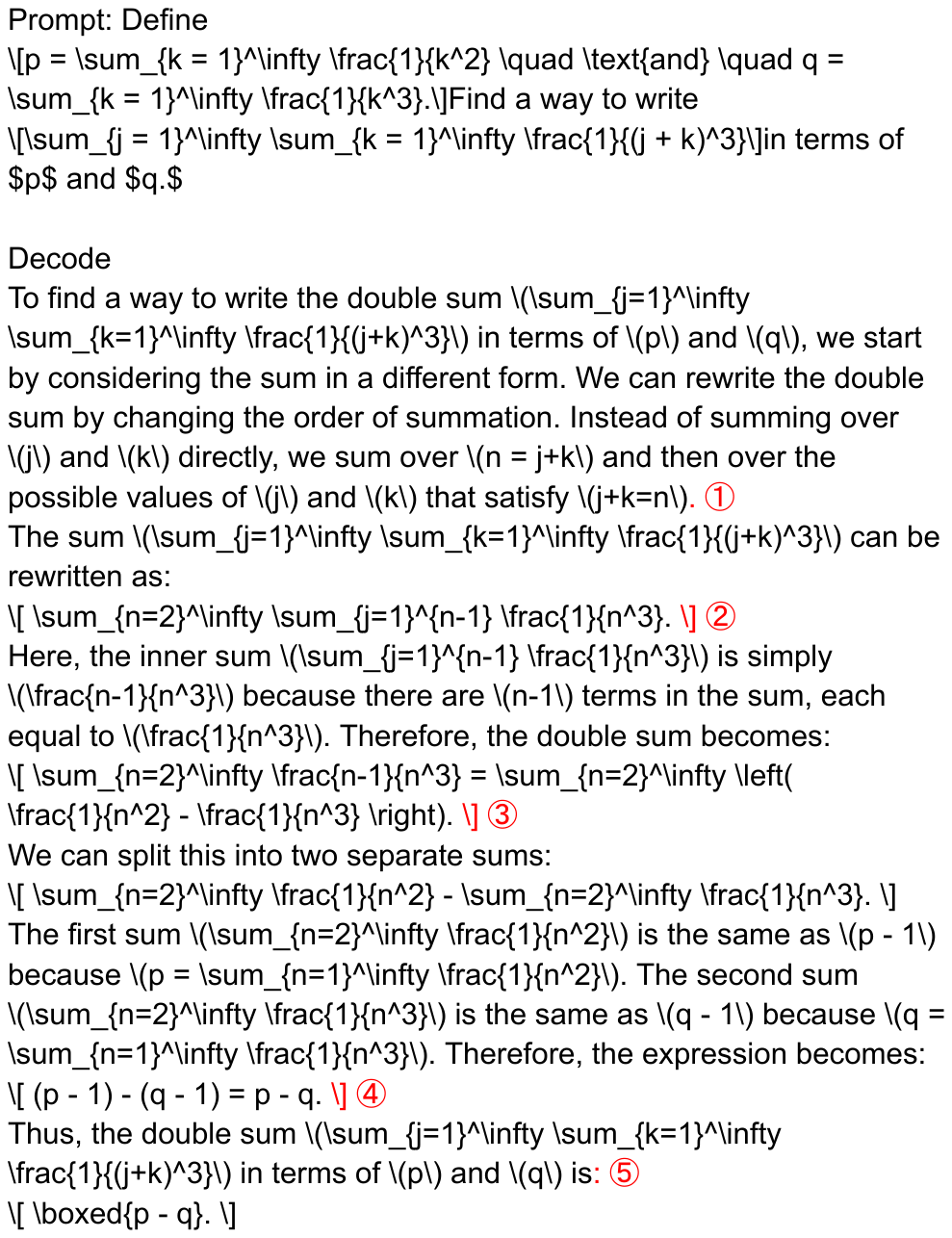}}}
\caption{Milestone example.}
\label{fig-appendix-milestone1}
\end{center}

\end{figure}

\section{More Examples for Milestone Tokens}

\begin{figure}[th]

\begin{center}
\fbox{\centerline{\includegraphics[width=\columnwidth]{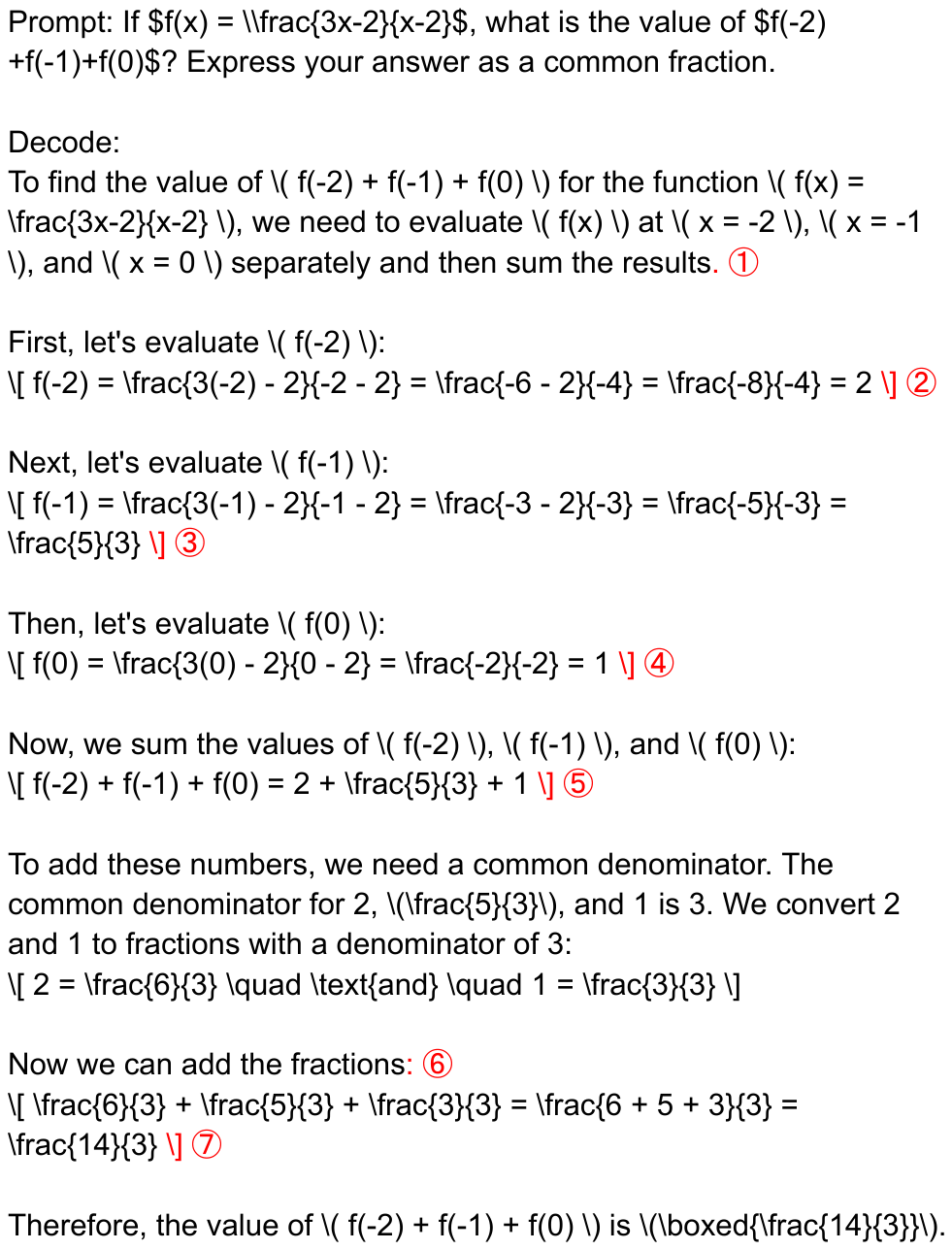}}}
\caption{Milestone example.}
\label{fig-appendix-milestone2}
\end{center}

\end{figure}

This section presents more examples of milestone tokens in Figure~\ref{fig-appendix-milestone1} and Figure~\ref{fig-appendix-milestone2}. For all examples, we input the prefill tokens to Qwen2.5-Math-7B-Instruct and obtain the corresponding decode tokens, as shown in the figure. The tokens marked red represent the milestone tokens. Although we only show a few examples here, the milestone patterns abound in reasoning tasks.

\section{Checklist-Related Issues}

Three datasets GSM8k (MIT), MATH500 (MIT), AIME (MIT), and four models Mistral Math 7B (No licence), Qwen 2.5 Math 7B Instruct (apache-2.0), Marco o1 (apache-2.0), DeepScaleR 1.5B Preview  (MIT) are used with their intended usage scenarios. We retrieve all models and datasets from Hugging Face, where detailed documentation, including parameter sizes and model architectures, is provided. We manually checked the data and believe there is no personal information misused.

We used ChatGPT to check the grammar of the texts.

To the best of our knowledge, we believe our work does not pose risks that harm any subgroup of our society.

\end{document}